\documentclass{article}
\usepackage{spconf,amsmath,graphicx}
\usepackage{microtype}
\usepackage{graphicx}
\usepackage{subcaption}
\usepackage{soul}
\usepackage{algorithm}
\usepackage{algpseudocode}

\usepackage{booktabs} 
\usepackage[dvipsnames,table]{xcolor}

\title{ Machine-Assisted Annotation of Forensic Imagery }
\newcommand{\method}{Proposed Annotations}

\newcommand{\dsname}{ITS-HD}
\newcommand{\size}{224}
\newcommand{\fmodel}{Model1}
\newcommand{\smodel}{Model2}
\name{Sara Mousavi$^{\star}$ \qquad Ramin Nabati$^{\star}$ \qquad Megan Kleeschulte$^{\dagger}$ \qquad Audris Mockus$^{\star}$}

\address{$^{\star}$  Department of Electrical Engineering and Computer Science\\
    $^{\dagger}$Department of Anthropology\\
    The University of Tennessee Knoxville, USA}

\begin{document}
%
\maketitle
\begin{abstract}
Image collections, if critical aspects of image content are exposed, can spur research and practical applications in many domains. Supervised machine learning may be the only feasible way 
to annotate very large collections, but leading approaches rely on large samples of 
completely and accurately annotated images. In the case of a large forensic collection, we are aiming to annotate, neither the complete annotation nor the large training samples can be feasibly produced. We, therefore, investigate ways to assist manual annotation efforts done by forensic experts. We present a method that can propose both images and areas within an image likely to contain desired classes. Evaluation of the method with human annotators showed highly accurate classification that was strongly helped by transfer learning. The segmentation precision (mAP) was improved by adding a separate class capturing background, but that did not affect the recall (mAR). Further work is needed to both 
increase the accuracy of segmentation and enhances prediction with additional covariates affecting decomposition. We hope this effort to be of help in other domains that require weak segmentation and have limited availability of qualified annotators. 
\end{abstract}
\vspace{-.05in}
\vspace{-.15in}
\section{Introduction}
\label{intro}
\vspace{-.1in}
Certain image collections, such as images of human decomposition, represent high potential value 
to forensic research and law enforcement, yet are scarce, have restricted access, and are very difficult to utilize. 
To make such image collections suitable for research and for law enforcement, they need to be annotated with relevant forensic 
classes so that a user can find images with the desired content. This work is motivated by our attempt to annotate over one million photos taken over seven years in a facility focused on studying human decomposition. 

Annotating images is a difficult task in general, with a single image taking from 19 minutes \cite{caesar2016coco} to 1.5 hours \cite{Cordts_2016_CVPR} on average. 
Human decomposition  images present additional difficulties. First, forensic data cannot be crowd-sourced due to its graphic nature and need for anonymity. 
Second, annotating forensic classes 
requires experts in human decomposition who are hard to come by. Therefore, it is natural to consider approaches 
to support such manual effort with machine learning (ML) techniques~\cite{andriluka2018fluid}. Unique challenges 
specific to forensic images prevent direct application of state-of-the art techniques described in, for example,~\cite{andriluka2018fluid}. This is mainly due to the primary focus of the annotation in being used by researchers, not algorithms. Particularly, when it comes to creating relevant training samples for ML approaches, we encountered the following challenges: 
\vspace{-.1in}
\begin{itemize}
    \item It is not feasible to annotate images completely. In other words, the user may choose to only annotate some instances of a class in an image, or only a subset of classes.
\vspace{-.1in}
    \item The locations of forensically-relevant objects is not precisely outlined but, instead, roughly indicated via rectangular areas.
\vspace{-.1in}
    \item It is not feasible to annotate a very large number of examples of a forensic class.
\vspace{-.1in}
\end{itemize}
The first challenge results from the numerous instances of certain classes (for example, there may be tens of thousands of maggots in a single image spread in multiple groups). Annotator may tag only classes relevant to annotator's investigation or classes that they have sufficient expertise to identify accurately. 
The second challenge is caused by the primary objective of the annotator to provide indicators to other researchers and the need to maximize the number of manually annotated images irrespective of the ability of machine learning to generalize from them (i.e. using simple rectangles instead of more time-consuming masks).
The last challenge is imposed by the limited availability of forensic experts. Furthermore, since it is not possible to annotate the entire set of images, the expert needs to choose which images to annotate.
Choosing images randomly, as it turns out, is highly inefficient since such images rarely contain relevant forensic classes.

LabelMe~\cite{russell2008labelme} and similar polygon-drawing interfaces have been used to annotate image collections~\cite{Cordts_2016_CVPR, mottaghi2014role, xiao2016sun, zhou2017scene}. The annotators need to manually select the areas of interest and label them with the correct label. Given the amount of time needed to annotate a single image, such approaches are not suitable for annotating one million forensic images. 

Fluid Annotation~\cite{andriluka2018fluid} assists annotators in fully annotating images by providing initial annotations, that can be edited as needed. Fluid annotation uses Mask-RCNN \cite{he2017mask} as the primary deep learning model. For Mask-RCNN  and other deep-learning based techniques such as Deeplabv3+ and YOLO~\cite{chen2018deeplab,redmon2017yolo9000} to work, large, complete, and clean training datasets such as Open Images, Image Net and COCO~\cite{kuznetsova2018open, deng2009imagenet, caesar2016coco} are required. 
Such approaches without additional training do not work for a dataset with a complete different set of object classes. Our attempts to train Mask-RCNN on the photos 
of human decomposition had extremely poor performance (even with transfer learning) due to incomplete set of annotations, approximate bounding boxes, and relatively few labeled instances per class. 

Other approaches to reduce the annotation effort involve using weakly annotated data, image-level or object- level annotations, 
for object detection~\cite{bilen2016weakly, gokberk2014multi, deselaers2010localizing} and semantic segmentation~\cite{kolesnikov2016seed, pathak2015constrained, khoreva2017simple, maninis2018deep}.
Although these approaches have been successful to some extent, there is 
still a large performance gap between the models trained on full segmentation masks and 
ones trained on image-level or object-level labels.

The main goal of this work is to simplify and speed-up the annotation process of forensic imagery by developing a machine-assisted semantic segmentation system (called~\method) that recommends potential annotations to annotators to simply accept or decline. 

\begin{figure}
    \centering
    \includegraphics[width=.7\columnwidth]{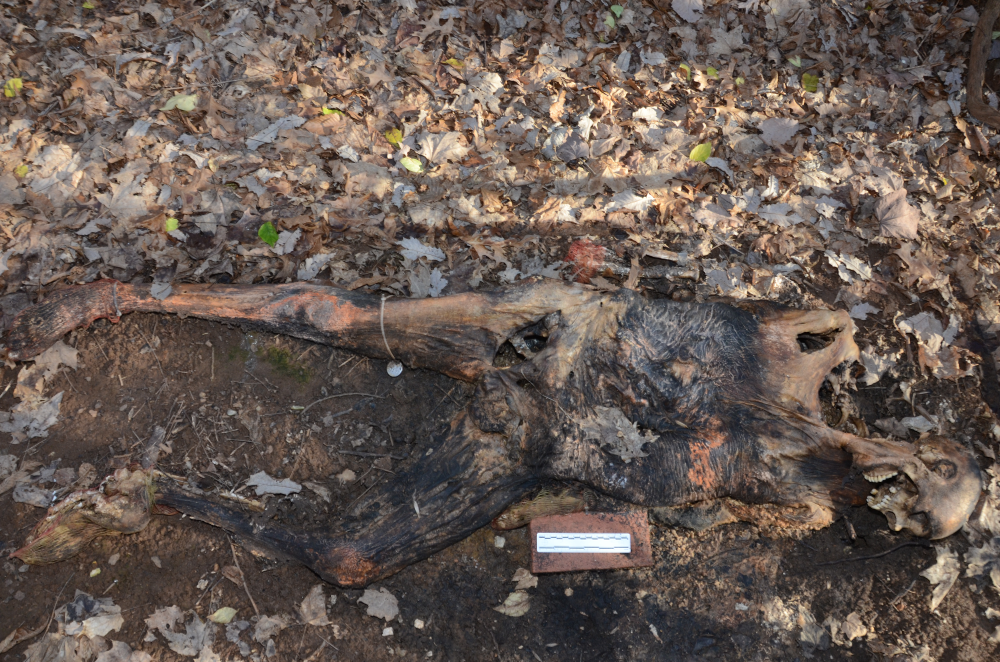}
    \caption{A sample image from \dsname. The image highlights the texture-like nature of the data. }
    \label{fig:img_example}
\vspace{-.1in}
\end{figure}

Semantic segmentation need a large training set. Our technique relies on the fact that human decomposition 
images are dominated by texture-like patterns (Figure \ref{fig:img_example}) repeated throughout a class. Our method, therefore, 
can work with a simpler classifier and less data. It utilizes the classifier in combination with a region selection technique to produce potential annotations provided to expert annotators. 


In addition, our approach can be used to estimate probabilities of a specific forensic class being present in the un-annotated set of images. While this is possible with other semantic segmentation methods, it is of particular use in forensic data, where a major problem faces the annotator: how to design sampling strategies to select images for manual annotation from the collection of one million images. 

\par Therefore, our contribution in this work is twofold. First, we present a novel semantic segmentation technique using a classifier and a region selector for forensic data, leveraging their pattern-like nature. Second, we use this method to propose not only new regions of interest for annotation, but also new images that are likely to contain classes of interest. 

The rest of the paper is as follows. Section~\ref{method} details our method and implementation. Section~\ref{result} discusses the results of our work and the paper is concluded in Section~\ref{conclusion}. 
\vspace{-.2in}
\section{\method}
\label{method}
\vspace{-.1in}
   \begin{figure*}
        \begin{minipage}[c]{0.48\textwidth}
            \centering
            \noindent\includegraphics[width=\textwidth,height=3cm,trim=4 4 4 4,clip]{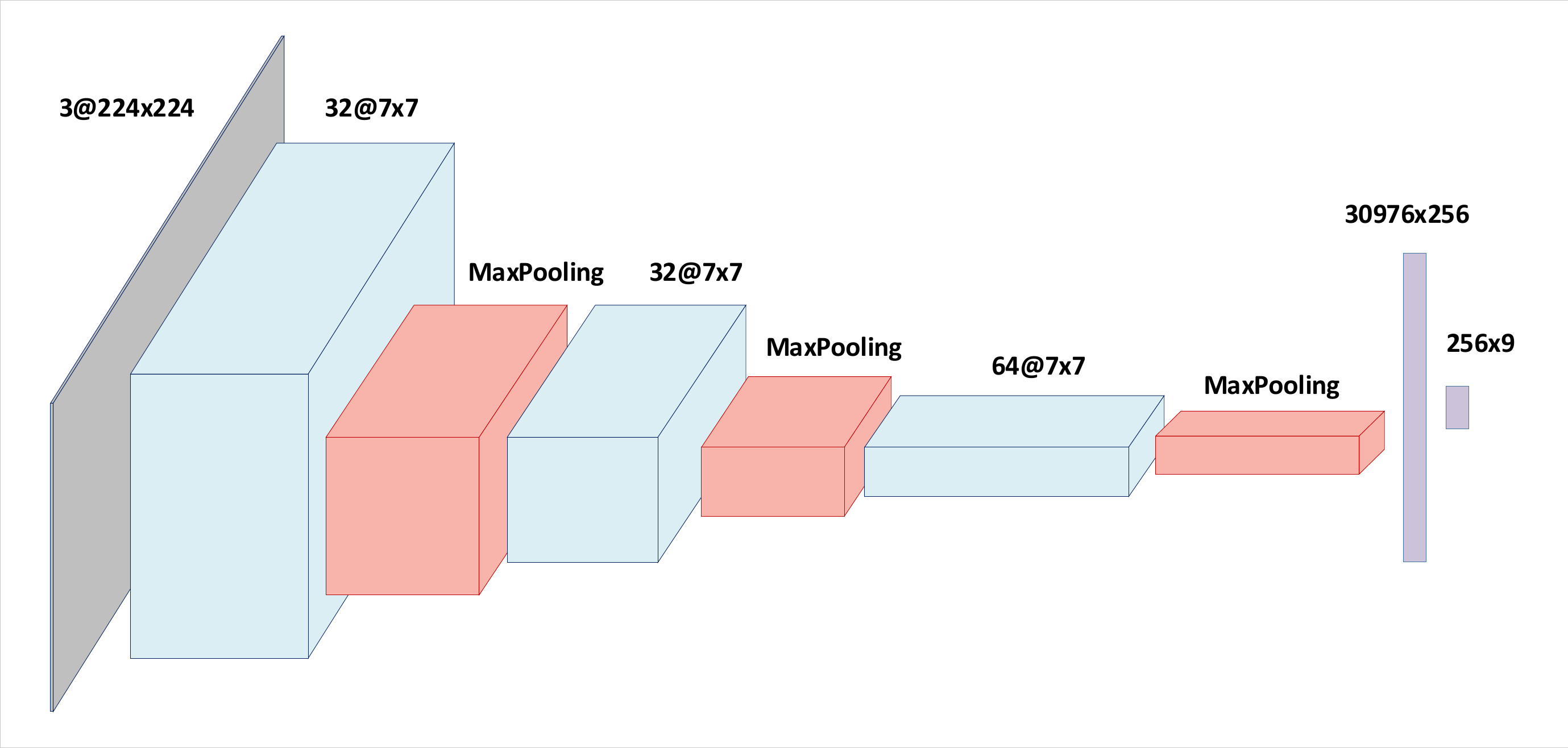}
            (a)
        \end{minipage}
        \begin{minipage}[c]{0.48\textwidth}
            \centering
            \noindent\includegraphics[width=\textwidth,height=3cm,trim=4 4 4 4,clip]{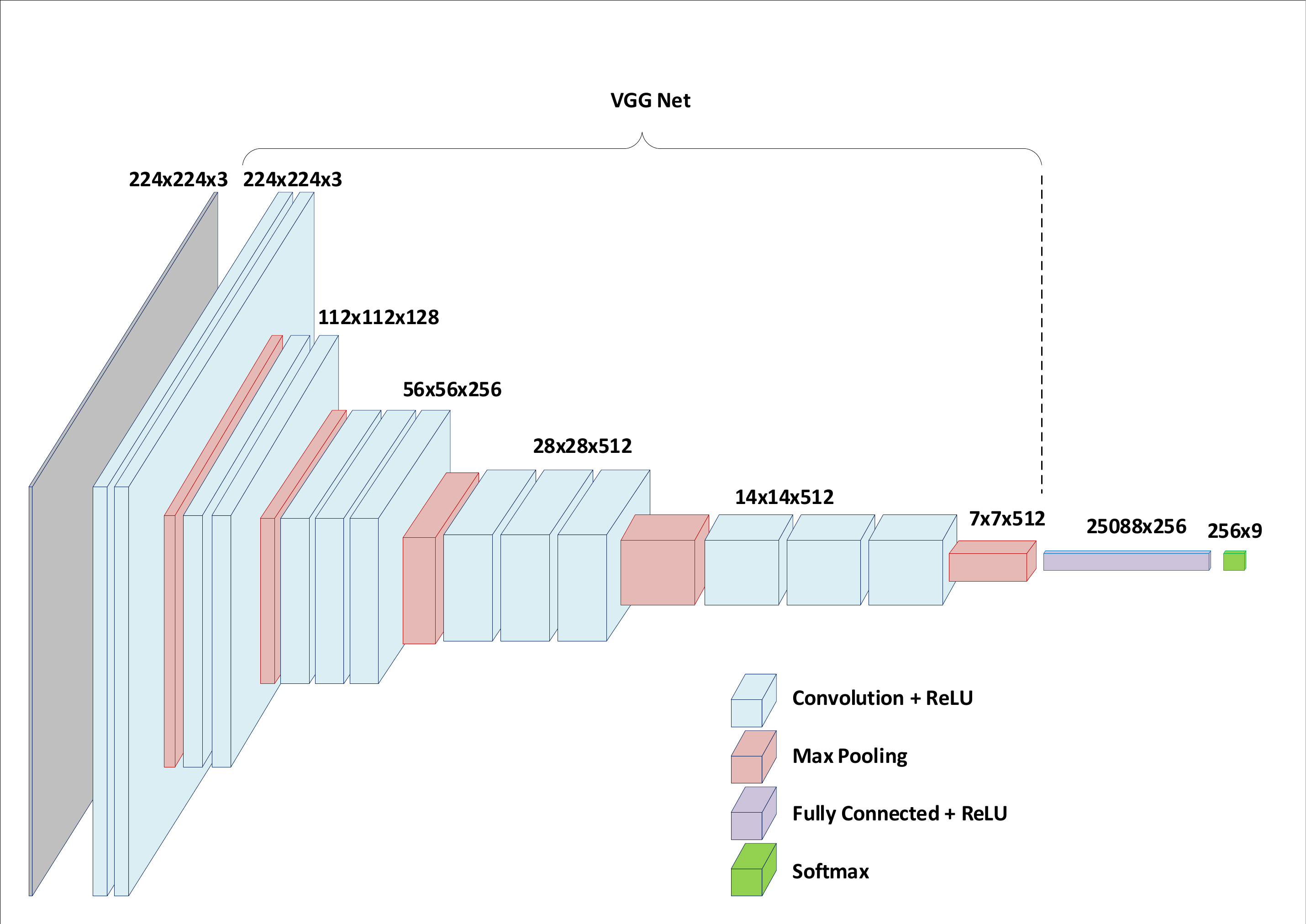}
            (b)
        \end{minipage}
\vspace{-.1in}
    \caption{The architecture of \fmodel~(a), and the VGG-based \smodel~(b) is shown. }
    \label{fig:arc}
\vspace{-.2in}
    \end{figure*}
\method~is comprised of a classifier and a region selection method. The classifier is trained on images that contain a single class. It is then used along with a region selection method to detect regions of new images. The classified regions are then systematically merged into larger segments resulting in semantic segmentation.
This process has three main steps: data preparation (Section~\ref{dataprep}), classification (Section~\ref{classification}) and semantic segmentation (Section~\ref{seg}). 
\vspace{-.1in}
\subsection{Human Decomposition Dataset} 
    \label{dataset}
\vspace{-.1in}
    Our image collection includes photos that depict decomposing corpses donated to the Anthropology Research Facility of our university. The photos are taken periodically from various angles to show the different stages of body decomposition. The collection spans from 2011 to 2016, and has over one million images. We call this dataset \dsname: Images Tracking Stages of Human Decomposition.
    \par The annotation for a small subset of this dataset has been done manually by four forensic experts resulting in 2865 annotated images. 
    However, as previously mentioned, these images are not fully annotated. 
    \par A sample image from \dsname~is shown in Figure \ref{fig:img_example}. The cadaver is mostly camouflaged in the background patterns.
    \par The forensic classes used in this work along with the number of annotated instances for each, are shown in Table~\ref{tbl:classes}.
    \vspace{-.3in}
    \subsubsection{Manual annotation}
    \label{annotation-protocol}
    \vspace{-.1in}
        To enable manual annotation of the small subset, we built an online platform that allows browsing, querying, and annotating \dsname. The online platform allows an authenticated user to log-in and annotate images. Since annotators sometimes use varying terms to identify the same forensic classes, we first decided on a nomenclature of terms. All annotators then used the developed nomenclature to label forensic classes in images. 
        To annotate a region in an image, the annotator first selects a rectangular bounding box around the region of interest and then enters the appropriate class name in an input dialog. The bounding boxes' coordinates along with the class names are then stored in a database. 
        \begin{table}[t]
\vspace{-.05in}
            \caption{An overview of the forensic classes of ITS-HD used in this work. The number of annotated instances is shown.}
            \label{tbl:classes}
            \vspace{-0.15in}
            \begin{center}
                \begin{footnotesize}
                    \begin{sc}
                        \begin{tabular}{p{2.3cm}cp{2.cm}c}
                            \toprule
                            Class name & \#samples &Class name & \#samples  \\
                            \midrule
                            maggots&1375&eggs&533\\
                            scale&716&mold&339\\
                            purge&709&marbling&241\\
                            mummification&557&plastic&107\\
                            \bottomrule
                        \end{tabular}
                    \end{sc}
                \end{footnotesize}
            \end{center}
\vspace{-.2in}
        \end{table}
\vspace{-.2in}
    \subsection{Data Preparation}
    \label{dataprep}
\vspace{-.1in}
        Preparing training data is a crucial step for making a highly accurate classifier. 
        Due to the similarity of some forensic classes to the background, both in terms of color and texture,  we added an additional class to the actual forensic classes for ``background''. 
        We then extracted areas designated as the forensic classes from the annotated images and used the class name to label each cropped section. Therefore, each annotation became a new training image by itself. 
        For the images extracted for ``background'', in order to create a diverse range of training data, we used a sliding window to extract smaller images from each training image.
        We re-sized all images to \size*\size~ and, as is commonly done, we also generated additional training data from the existing annotations using data augmentation such as flip, zoom, scale, and shear.
       \vspace{-.1in}
    \subsection{Classification}
    \label{classification}
    \vspace{-.1in}
         We used a CNN with a multinomial logistic regression classifier to train a model for classifying regions of the un-annotated images.
         The preponderance of texture-rich classes did not call for very deep neural networks. 
        We started with \fmodel~that uses a simple neural network shown in Figure \ref{fig:arc}:a. The CNN network in this model has three convolutional and two fully connected layers. We used normalization after each layer and also a drop-out of 0.5 before the last layer. In addition to \fmodel,~we also experimented with \smodel~, a standard VGG with two fully connected layers added on top. Images generated from section \ref{dataprep} were used to train and validate these two models. We trained \fmodel~from scratch. However for \smodel, we tested both pre-trained weights obtained from ImageNet as well as random weights. 
    \vspace{-.1in}
    \subsection{Semantic Segmentation}
    \label{seg}
    \vspace{-.1in}
        Locating the forensic objects within images is done using the classifier described in section \ref{classification}. Algorithm \ref{alg:seg} shows how semantic segmentation is done in \method. Regions of un-annotated images are fed into the classifier model to be classified. The regions are generated using a sliding window of size \size*\size~with a stride of 200. Since the training data is not fully annotated, many regions within an image may contain classes that the classifier has not been trained on. 
        To reduce the number of false positives, we use a threshold of 0.85  to accept a classification done on a region, otherwise it will be ignored. To generate the regions, a sliding window is slid over the entire un-annotated image. 
        \par A brute-force sliding window method for selecting regions is often used with different window sizes.  Since we focus on detecting patterns at roughly similar scales, varying window sizes is not necessary. 
        \par The contiguous classified regions of the images need to be organized so that neighbor regions belonging to the same class are proposed as a single composite segment. To do so, we group the classified regions by first finding overlaps. Then, we create an adjacency matrix of size $n \times n$ where $n$ is the number of regions for the class. A cell $(i, j)$ (for two regions $i$ and $j$) is set to 1 if the two regions overlap. We then create a graph from the adjacency matrix and find the connected components of the graph for each class using the igraph library \cite{igraph}. Next, we find the convex hull for each connected component. The resulting hulls are presented to the annotator as proposed annotations. The confidence of a recommended annotation is calculated based on the average confidence of the individual regions within that component.   
    \begin{algorithm}
    \footnotesize
    \caption{Semantic segmentation in our method}
    \label{alg:seg}
    \begin{algorithmic}[1]
    \Procedure{segment}{image}
    \For{every $region$ in image}
    \State \textsc{Classify}($region$)
    \State Store $region$'s coordinate, class\_id and confidence
    \EndFor
    \For{every $c$ in $classes$}
    \State Find all regions classified in class $c$
    \State Create the adjacency matrix
    \State Group neighbor regions in one single area
    \State Calculate score for each area
    \EndFor
    \State Present the segmentation as proposals to the annotator
    \EndProcedure
    \end{algorithmic}
    \end{algorithm}
\begin{figure*}[t!]
        \centering
        \begin{minipage}[c]{0.20\textwidth}
            \centering
            \noindent\includegraphics[width=\textwidth]{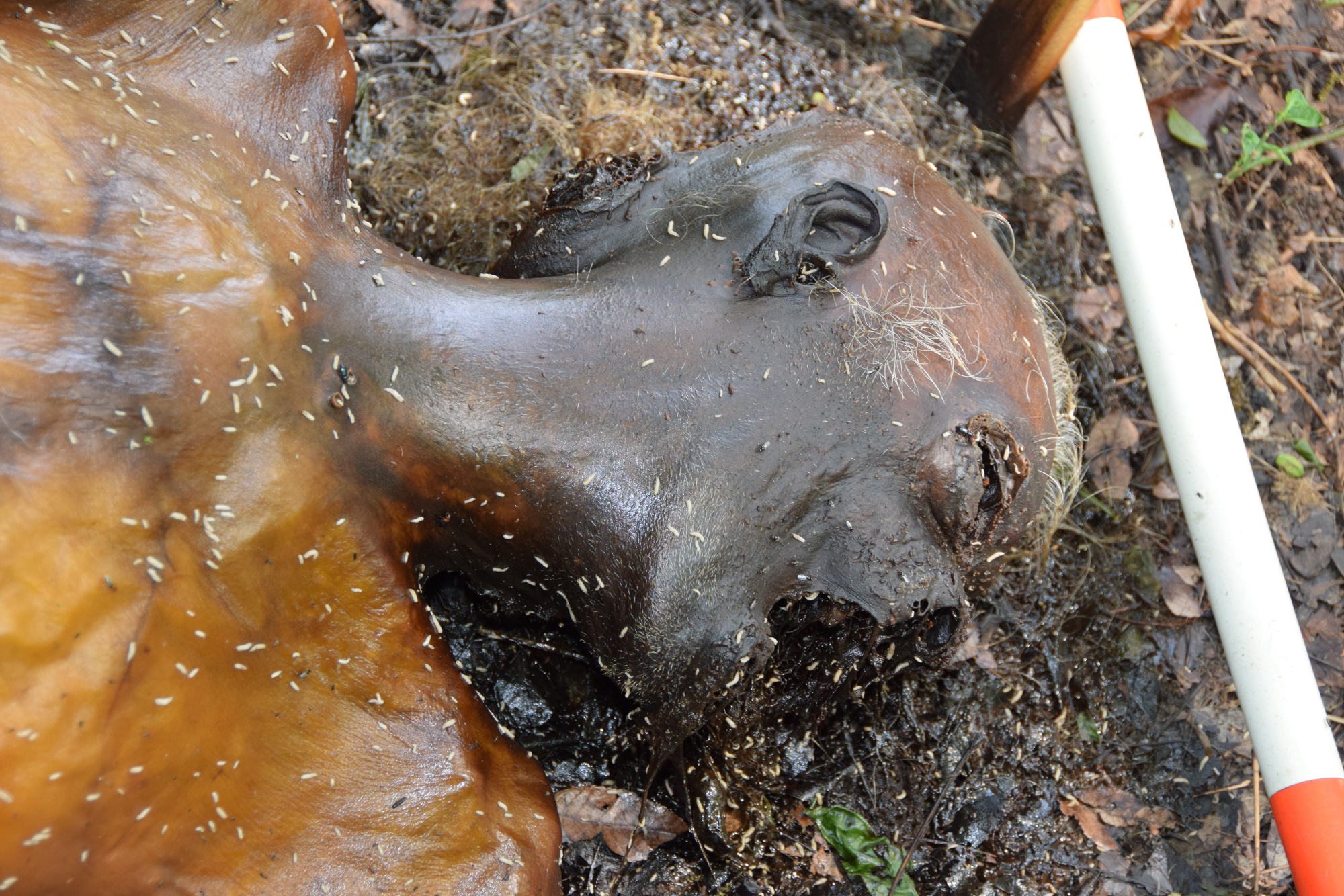}
            (a)
        \end{minipage}
        \begin{minipage}[c]{0.20\textwidth}
            \centering
            \noindent\includegraphics[width=\textwidth]{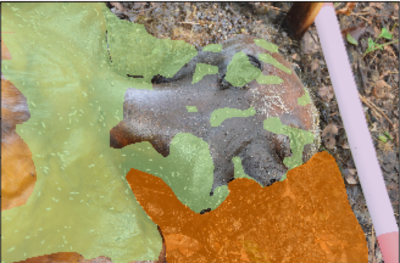}
            (b)
        \end{minipage}
        \begin{minipage}[c]{0.20\textwidth}
            \centering
            \noindent\includegraphics[width=\textwidth]{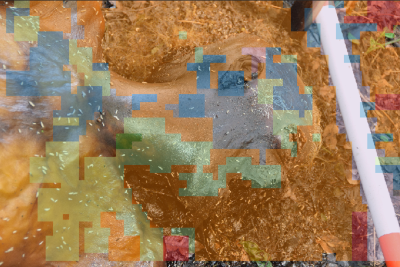}
            (c)
        \end{minipage}
        \begin{minipage}[c]{0.20\textwidth}
            \centering
            \noindent\includegraphics[width=\textwidth]{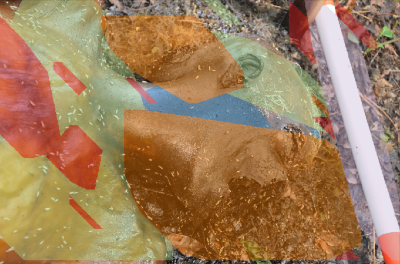}
            (d)
        \end{minipage}
        \begin{minipage}[c]{0.10\textwidth}
            \centering
            \noindent\includegraphics[width=\textwidth]{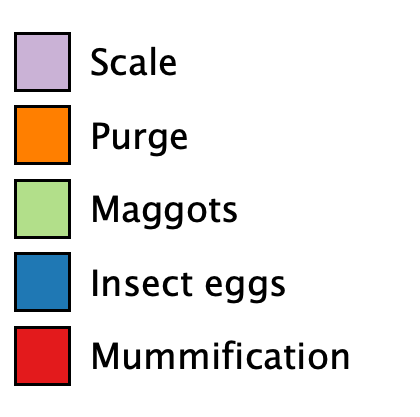}
            
        \end{minipage}
        \vspace{-.1in}
        \caption{Detected forensic classes using \fmodel-bg and \smodel-bg-tl are shown in (c) and (d) respectively. Sub-figure (a), (b) show the original image and the ground truth respectively. }
        \label{fig:results}
\vspace{-.2in}
\end{figure*}
\vspace{-.3in}
\section{Results and discussion}
\label{result}
\vspace{-.1in}
To evaluate \method, we measure the accuracy in comparison to the manual annotation done by a forensic expert. The results include the performance of \fmodel~and \smodel. Additionally, we tested the effects of including the background as a separate class in both models and also the effect of transfer learning on \smodel. 
    Section \ref{setup} describes tuning parameters for both models and evaluation setup. Section \ref{metric} discusses our findings. 
    \vspace{-.1in}
    \subsection{Evaluation Setup}
    \label{setup}
    \vspace{-.1in}
        \method~is implemented using Keras, TensorFlow and Python. We used MongoDB as our database. 
        For both CNN networks we used the $SGD$ optimizer with a learning rate of $0.001$.
        
        \par Over two hundred distinct classes of samples were present in the dataset. To select a more manageable number of classes for the experiments, we first excluded classes with fewer than 100 ground truth instances and asked forensic experts to select the most important classes for the forensic community.
        We used one third of images per class for validation and the remaining images for training.
    
        \par To evaluate the performance of our \method, we randomly selected 46 images and asked a forensic expert to provide us with the ground truth annotation masks only for the forensic classes used in this work. These images were annotated carefully and completely with polygonal selections, taking about 3 hours to complete. We evaluated the performance of our proposed annotations against these round truths.
   \vspace{-.1in}
    \subsection{Discussion}
    \label{metric}
    \vspace{-.01in}
    Table~\ref{table:results} shows the performance of \method. We calculated mean average precision (mAP) for the classification done by both \fmodel~and \smodel~over all classes. We also calculated mean average recall and precision over all classes (mAR, mAP) for our semantic segmentation against the ground truth. These values are used as mAP and mAR in Table \ref{table:results}. 
    
    The mean average precision is calculated as the ratio of correct predicted pixels over the total predicted pixels for each class. This value is then averaged for each class over all 46 images. We used a similar method for mAR, however we used the the ratio of correct predicted pixels to the total ground truth pixels for each class. 
    
    Table~\ref{table:results} shows that transfer learning improves the performance of \smodel. Comparing \smodel~with \smodel-tl, we can see that transfer learning has improved both mAP of the classifier model and mAR of the semantic segmentation.
    
    Comparing \smodel~with \fmodel~in Table \ref{table:results}, we believe that we might get even better results using \fmodel~if we first train it on another dataset such as ImageNet, considering the fact that \fmodel~is a very simple model and its training takes less time compared to \smodel.
    
    A trade off between using a model with high recall or high precision can also be observed from the table. For the purpose of suggesting classes to a human annotator, it is more important to detect a forensic class if it exists, as opposed to exactly pinpointing the location of the class within the image. Thus, we want to have a model with higher recall and a reasonable $mAP$. Our results also show that including the background as a class improves mAP for the semantic segmentation. 
    
    Figure \ref{fig:results} shows a segmentation using \fmodel~without transfer learning and \smodel~with transfer learning, and compares it to the ground truth. Both models were trained on 8 forensic classes plus the background class. We can see that a better segmentation is obtained when transfer learning is employed. 

    \begin{table}[t]
        \caption{Performance of classifier models and semantic segmentation in \method. bg and tl stand for background and transfer learning.}
        \centering
        \begin{footnotesize}
        \begin{tabular}{c | c c | c}
        \hline\hline
         & \multicolumn{2}{c}{Semantic Segmentation} & \multicolumn{1}{c}{Classification}\\
        Method & mAP & mAR & mAP\\ [0.5ex]
        \hline
        \smodel-bg-tl & 0.26 & 0.45 & 0.95\\ 
        \smodel-tl & 0.15 & 0.59 & 0.92\\
        \smodel & 0.30 & 0.28 & 0.79\\
        \fmodel & 0.16 & 0.32 & 0.84 \\
        \fmodel-bg & 0.17 & 0.23 & 0.88 \\
        \hline
        \end{tabular}
        \label{table:results}
        \end{footnotesize}
    \end{table}

\section{Conclusion}
\label{conclusion}
\vspace{-.01in}
In this work, we discuss an annotation-assistance system that proposes annotations within an image as well as images likely to contain a desired class to forensic experts. At the core of our system we introduce a semantic segmentation method composed of 
a classifier in conjunction with a sliding-window-based region selection method. We also evaluate its applicability 
in the context of imagery documenting human decomposition where classes are primarily determined by patterns. 
We demonstrate the feasibility of semantic segmentation in this domain using a relatively small set of training samples. 
As is expected with small training samples, transfer learning has brought noticeable improvements in accuracy. 
Inclusion of the background as a class also brought improvements, possibly because background is at times difficult 
to distinguish from focal classes. 

In the future, we would like to evaluate if our method would work with other types of texture-like data. 
In addition, we plan to utilizing body pose detection methods to improve the ability to exclude background
and increase the accuracy of our system for forensic class segmentation.

\bibliographystyle{IEEEbib}
\bibliography{refs}

\end{document}